\documentclass[pdflatex,sn-mathphys-num]{sn-jnl}

\usepackage{graphicx}%
\usepackage{multirow}%
\usepackage{amsmath,amssymb,amsfonts}%
\usepackage{amsthm}%
\usepackage{mathrsfs}%
\usepackage[title]{appendix}%
\usepackage{xcolor}%
\usepackage{textcomp}%
\usepackage{manyfoot}%
\usepackage{booktabs}%
\usepackage{algorithm}%
\usepackage{algorithmicx}%
\usepackage{algpseudocode}%
\usepackage{listings}%
\usepackage{array}%
\usepackage[table]{xcolor}

\definecolor{myred}{HTML}{CA0000}
\definecolor{mygreen}{HTML}{008732}
\definecolor{myblue}{HTML}{0066CB}

\usepackage{multirow}   
\usepackage{colortbl}
\definecolor{gray}{gray}{0.5}
\definecolor{c2}{HTML}{FBD9BD}
\definecolor{c3}{HTML}{fe793d} 
\definecolor{c4}{HTML}{eedeb0}
\definecolor{rouse}{rgb}{0.981,0.961,0.941}

\usepackage{algorithm}
\usepackage{algpseudocode}
\algrenewcommand\algorithmicrequire{\textbf{Inputs:}}
\algrenewcommand\algorithmicensure{\textbf{Outputs:}}

\usepackage{caption}

\usepackage{subcaption}

\makeatletter
\@ifundefined{cline}{\newcommand{\cline}[1]{\cmidrule(lr){#1}}}{\renewcommand{\cline}[1]{\cmidrule(lr){#1}}}
\makeatother

\theoremstyle{thmstyleone}%
\theoremstyle{thmstyletwo}%

\theoremstyle{thmstylethree}%

\begin{document}

\title[Article Title]{CamoNAS: Neural Architecture Search for Enhanced Camouflaged Object Detection}


\author*[1,2]{\fnm{Dawei} \sur{Ren}}\email{rendawei23@mails.ucas.edu.cn}

\author[1,2]{\fnm{Yan} \sur{Zhang}}\email{zhangyan232@mails.ucas.ac.cn}

\author[1,2]{\fnm{Hongying} \sur{Tang}}\email{tanghy@mail.sim.ac.cn}

\author[1,3]{\fnm{Qiaoling} \sur{Zhou}}\email{zhouql2023@shanghaitech.edu.cn}

\author[1,2]{\fnm{Jianpo} \sur{Liu}}\email{liujp@mail.sim.ac.cn}

\affil*[1]{\orgdiv{Science and Technology on Micro-system Laboratory}, \orgname{Shanghai Institute of Microsystem and Information Technology, Chinese Academy of Sciences}, \orgaddress{\street{865 Changning Road}, \city{Shanghai},  \postcode{200050}, \country{China}}%
}

\affil[2]{%
	\orgname{University of Chinese Academy of Sciences}, %
	\orgaddress{\street{No.~19(A) Yuquan Road, Shijingshan District}, %
		\city{Beijing}, %
		\postcode{100049}, %
		\state{Beijing}, %
		\country{China}}%
}

\affil[3]{\orgdiv{School of Information Science and Technology}, \orgname{Shanghaitech University}, \orgaddress{\city{Shanghai}, \country{China}}}


\abstract{Camouflaged Object Detection (COD) aims to locate and segment objects that blend into their surroundings, presenting challenges due to weak edge cues and ill-defined boundaries. Traditional COD models rely on hand-designed architectures and multi-scale feature fusion, which are often guided by intuition rather than systematic search. This paper introduces CamoNAS, a frequency-aware multi-resolution Neural Architecture Search (NAS) framework for COD. CamoNAS automatically searches both cell-level operations and network-level downsampling paths, forming a hierarchical search space tailored to detect camouflaged objects. Additionally, it adopts an RGB frequency dual-stream architecture, where a learnable wavelet transform complements the RGB spatial stream. CamoNAS achieves state-of-the-art performance on four COD benchmarks (CAMO, COD10K, NC4K, CHAMELEON), highlighting the effectiveness of NAS for COD. Our code is available at \url{https://github.com/rendaweiSIMIT/CamoNAS}.}

\keywords{Camouflaged Object Detection, Neural Architecture Search, Concealed Object Segmentation}



\maketitle

\section{Introduction}

\label{sec:intro}
Camouflaged Object Detection (COD) aims to locate and segment objects that blend into their surroundings~\cite{fan2021concealed}. It has applications in wildlife conservation~\cite{fan2020camouflaged,yang2021uncertainty}, medical imaging~\cite{fan2020pranet,zhang2020adaptive}, transparent object detection~\cite{mei2020don,xie2020segmenting}, etc. Compared with conventional detection or segmentation, COD is more challenging because camouflaged targets often mirror the background's texture, color, and shape, leaving weak edge/gradient cues and ill-defined object-background boundaries. This setting demands precise boundary modeling and strong suppression of background distractors. In recent years, researchers have proposed many architectures and feature-extraction strategies, e.g. mimic human vision~\cite{zhang2022preynet,mei2021camouflaged,pang2022zoom}, attention mechanisms~\cite{cheng2025enhancing,kowalski2025bi,liu2022boosting}, frequency-domain cues~\cite{zhong2022detecting,He2023Camouflaged}, and joint learning with auxiliary vision tasks~\cite{zhai2021mutual,sun2022boundary,zhaiexploring,lv2021simultaneously,jia2022segment,zhu2022can}. Collectively, these directions alleviate COD's difficulty and achieve notable success across diverse scenarios.

\begin{figure}[htbp]  
	\centering\includegraphics[width=8cm]{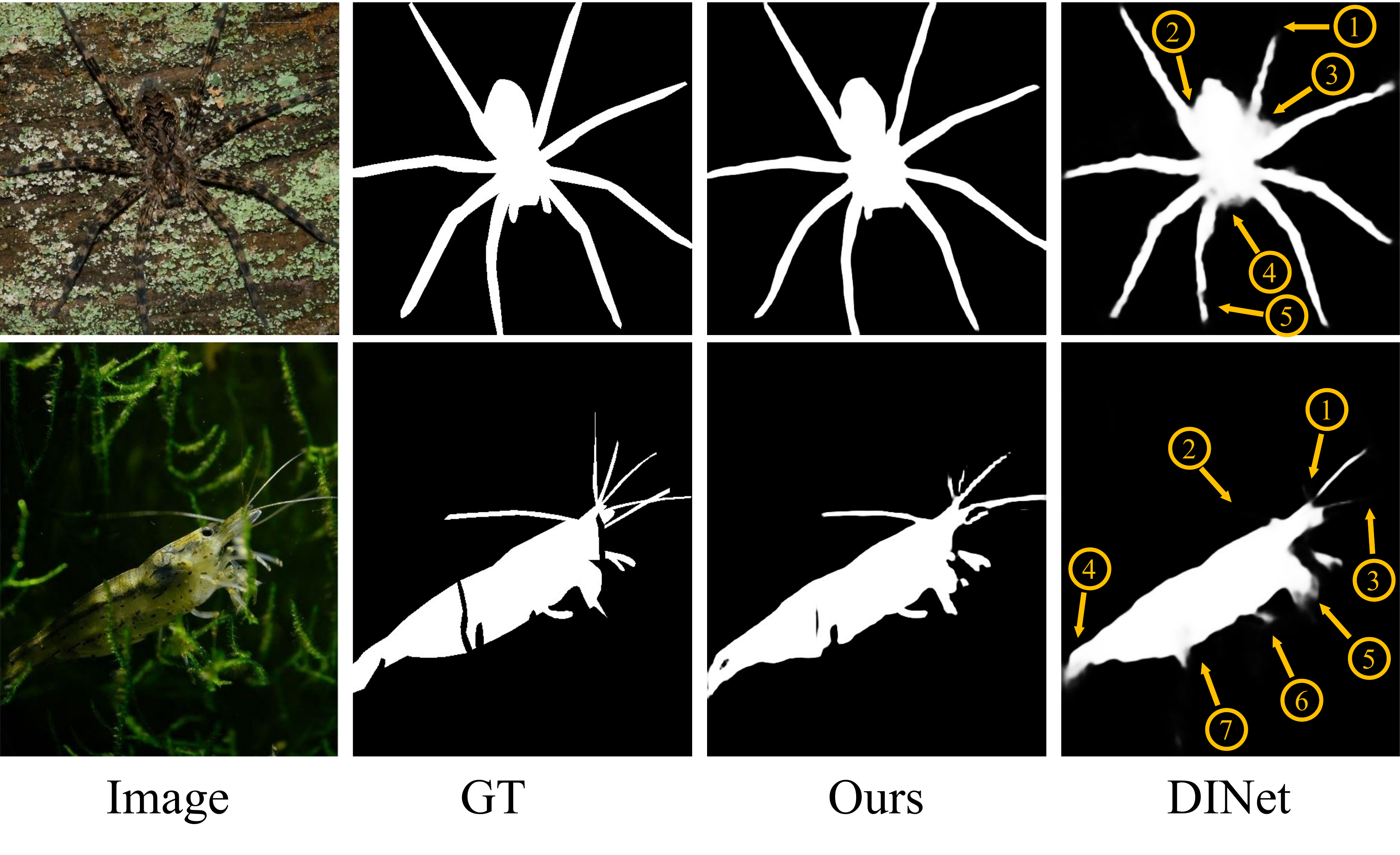} 
	\caption{Qualitative comparison on camouflaged object detection. Under severe foreground--background similarity, the state-of-the-art COD baseline DINet exhibits background leakage, blurred edges, and missing thin structures (orange arrows). Our CamoNAS suppresses these distractions and preserves fine details, producing cleaner masks and sharper contours.}
	\label{fig:cct}   
\end{figure}

However, most of the above methods rely on manually designed architectures, which require extensive trial-and-error with different network structures and hyperparameter combinations for each dataset or scenario. This process is not only time-consuming and labor-intensive, but it also risks getting stuck in local optima due to human bias, making it difficult to fully explore the vast space of possible model architectures. Neural Architecture Search (NAS)~\cite{elsken2019neural} is an automated model design paradigm that has achieved remarkable results in tasks such as image classification~\cite{zoph2016neural,zoph2018learning,liu2018progressive,real2019regularized,pham2018efficient} and semantic segmentation~\cite{liu2019auto,zhang2019customizable,nekrasov2019fast,lin2020graph}. NAS can automatically select network modules and connections from a large candidate search space~\cite{liu2018darts}, achieving an excellent trade-off between performance and model complexity. This capability provides a new perspective on how to design deep network structures. However, NAS for COD remains relatively underexplored. Recent work~\cite{li2025camouflaged} has demonstrated the feasibility of applying NAS to COD, while it remains open how to design COD-specific search spaces and strategies that jointly ensure precise boundary delineation and robust background suppression~\cite{liang2024systematic}. COD places stricter demands on precise boundary delineation and robust suppression of background distractions, leaving unresolved how to tailor the NAS search space and strategy to COD-specific requirements and how to automatically synthesize architectures truly suited to this task.

To address these challenges, we propose CamoNAS, a frequency-aware, multi-resolution NAS framework for COD. CamoNAS is designed to ease the burden of manual network design. Specifically, we build a hierarchical search space ranging from cell-level operator choices to network-level multi-scale path configurations. This design allows the model to dynamically adjust the feature-map resolution at each layer. High spatial resolutions capture fine edges and texture details, whereas low spatial resolutions encode abstract global semantics. In this way, the network satisfies both detail and semantic requirements for COD. Meanwhile, inspired by prior wavelet-like decomposition for COD (e.g., FEDER~\cite{He2023Camouflaged}) and the effectiveness of frequency-domain cues~\cite{cong2023frequency,xie2023frequency}, we introduce an LDWT that decomposes the input into four sub-bands (one low-frequency and three directional high-frequency maps) to form a parallel frequency stream alongside the RGB stream. With this RGB-frequency dual-stream architecture, the network can extract discriminative cues from both the spatial domain and the frequency domain separately before fusing them at a high semantic level. This design helps uncover subtle target textures and structural patterns that are hidden in complex backgrounds. In the decoding stage, we adopt a lightweight fusion head based on low-rank matrix decomposition~\cite{zhou2014low} to adaptively fuse the multi-scale features from both streams. This fusion module integrates cross-domain, multi-scale features while preserving key details, thereby reducing semantic conflicts and suppressing background noise. Collectively, these innovations enable CamoNAS to jointly achieve precise boundary delineation and accurate whole-object localization for camouflaged targets, thereby delivering significant improvements in segmentation accuracy and robustness under challenging camouflage.

In summary, our approach combines automated NAS-based modeling with COD-specific design considerations, producing high-performance COD models without tedious manual tuning. Our main contributions are as follows:

\begin{itemize}
	\item We propose a NAS framework tailored to COD, with a hierarchical search space spanning cell-level operators and network-level cross-scale routing, enabling end-to-end architecture optimization for camouflaged targets.
	
	\item We incorporate a learnable wavelet-based frequency decomposition module into NAS to form an RGB?frequency dual-stream network. The module produces four sub-bands (one low-frequency and three directional high-frequency components) and is trained with an explicit perfect-reconstruction regularization, providing complementary texture and structural cues for COD.
	
	\item We achieve competitive state-of-the-art performance on four benchmark COD datasets across all standard metrics, and comprehensive experiments, including ablation studies, confirm the effectiveness of our technique.
\end{itemize}

\section{Related Work}
\label{sec:Related_Work}

\subsection{Camouflaged Object Detection}

In COD tasks, the core difficulty is the high intrinsic similarity between foreground and background, which makes targets blend seamlessly into their surroundings~\cite{bi2021rethinking,mondal2017partially,li2018fusion,le2019anabranch}. To counter this, prior work enhances target cues while suppressing background noise. Representative designs include symmetric/multi-scale fusion (SINet)~\cite{fan2021concealed} for edge delineation, hard-region mining with residual fusion (PFNet)~\cite{mei2021camouflaged}, joint learning with saliency for contrast-aware localization, edge-guided supervision for complete structures (BGNet)~\cite{sun2022boundary}, and multi-scale zoom-in/zoom-out reasoning (ZoomNet)~\cite{pang2022zoom}. Frequency cues have also been introduced: learnable wavelet decomposition to expose textural signals, and masked separable attention to refine multi-level aggregation (CamoFormer)~\cite{yin2024camoformer}. 

Beyond these frequency-aware designs, recent COD models further improve performance by refining global--local fusion and iterative decoding. Yue \textit{et al.}~\cite{yue2025cnn} fuse convolutional and Vision Transformer representations at the decision level to couple global context with fine-grained details for camouflaged object segmentation, while Ge \textit{et al.}~\cite{ge2025feature} propose a feature-aware iterative refinement network that progressively recovers complete camouflaged objects via feature- and edge-aware refinement. These advances have improved results on \textit{CHAMELEON}~\cite{skurowski2018animal}, \textit{CAMO}~\cite{le2019anabranch}, \textit{COD10K}~\cite{fan2021concealed}, and \textit{NC4K}~\cite{lv2021simultaneously}. Yet most methods---including recent CNN--ViT hybrids and iterative refinement frameworks---rely on hand-crafted architectures and empirical tuning~\cite{fan2020pranet,fan2020camouflaged,lin2021rich}, limiting generalization across datasets and scenes. A recent survey~\cite{liang2024systematic} further highlights persistent gaps in fine detail reconstruction and robustness. This motivates automated architecture optimization, which is the direction we pursue.

Despite rich feature-extraction~\cite{ronneberger2015u,chen2017deeplab} and context-fusion strategies~\cite{kirillov2020pointrend,lin2017feature}, balancing fine-grained boundaries with global semantics remains challenging. Many designs employ deep fusion, attention, or auxiliary heads (e.g., boundary/saliency)~\cite{sun2022boundary,zhaiexploring,lv2021simultaneously,jia2022segment,zhu2022can}, often increasing complexity without consistent cross-domain generalization. Beyond COD, transformer architectures based on mutual attention for image anomaly detection~\cite{Zhang2023MutualAttention}, dual-branch Swin Transformer--ConvNeXt networks for strong-noise image denoising~\cite{Lin2025SCNet}, and robust segmentation under label noise for 3D point clouds~\cite{Zhang2025JointLearning} further demonstrate the importance of expressive yet robust multi-branch and multi-scale representations under challenging supervision. We therefore introduce NAS for COD to obtain end-to-end, data-driven architectures that reduce manual effort and improve robustness in varied camouflage scenarios.

\subsection{Neural Architecture Search for Semantic Segmentation}

NAS~\cite{elsken2019neural} automates topology and operator selection and has shown strong results in classification~\cite{zoph2016neural,zoph2018learning,liu2018progressive,real2019regularized,pham2018efficient,yao2020sm}, detection~\cite{zoph2018learning,chen2019detnas,guo2020hit,wang2020fcos}, and segmentation~\cite{liu2019auto,zhang2019customizable,nekrasov2019fast,lin2020graph}. Early RL/evolutionary NAS~\cite{zoph2016neural,real2017large} are computationally expensive; differentiable relaxation methods, pioneered by DARTS~\cite{liu2018darts} and its variants~\cite{xu2019pc,chen2019progressive,chu2020darts}, convert discrete architectural choices into learnable weights, thereby enabling efficient gradient-based search with significantly reduced computational cost. More recently, one-shot NAS frameworks such as MNGNAS distill adaptive combinations of multiple searched subnetworks into a single supernet, improving both search stability and final accuracy~\cite{Chen2023MNGNAS}. 

Extending NAS to dense prediction raises memory and multi-scale representation challenges~\cite{elsken2019neural}, but notable successes exist: NAS-FPN~\cite{ghiasi2019fpn} discovers superior feature pyramids, MnasNet~\cite{tan2019mnasnet} optimizes accuracy--latency for mobile via device-aware objectives, and EfficientDet~\cite{tan2020efficientdet} couples BiFPN with compound scaling for strong COCO performance. Other variants improve practicality---ProxylessNAS~\cite{cai2018proxylessnas} performs direct, target-task search without proxies, while progressive strategies~\cite{liu2018progressive} accelerate exploration. In parallel, efficient attention pyramid transformers, such as EAPT~\cite{Lin2023EAPT}, demonstrate that carefully designed multi-scale attention can benefit image classification, object detection, and semantic segmentation, reinforcing the importance of hierarchical global--local modeling within the search space. 

Collectively, these works show that NAS and attention-based backbones can find architectures that outperform hand-crafted baselines while balancing accuracy and complexity~\cite{liu2018darts,liu2019autodeeplab,cheng2020hierarchical,na2022autosnn}. Beyond general dense prediction, recent work has also brought NAS into COD. Li \textit{et al.} propose ALRNet~\cite{li2025camouflaged}, which searches camouflage-specific modules for coarse localization and edge-assisted refinement. Building on this line of research, we explore a complementary direction: CamoNAS jointly searches macro-level multi-resolution routing (down/hold/up) together with cell-level operators, and further integrates a learnable wavelet-based frequency stream into the NAS process to derive frequency-aware architectures for sharper boundaries and stronger generalization in camouflage scenes.

\section{Methodology}
\label{sec:Methodology}
\paragraph{Motivation.} Motivated by the limited design space and heavy tuning burden of hand-designed networks~\cite{elsken2019neural}, we pursue NAS-driven topology discovery for COD. We replace manual heuristics with an automatically searched topology: a differentiable, resolution-aware NAS that co-optimizes cell operators and down/hold/up paths. In parallel, a learnable discrete wavelet transform forms a frequency stream that complements the RGB stream, and a lightweight late-fusion head aggregates both. This design allocates capacity across sub-bands and resolutions, sharpening edges and suppressing distractors without heavy hyperparameter tuning.

\begin{figure*}
	\begin{center}
		\includegraphics[width=13cm]{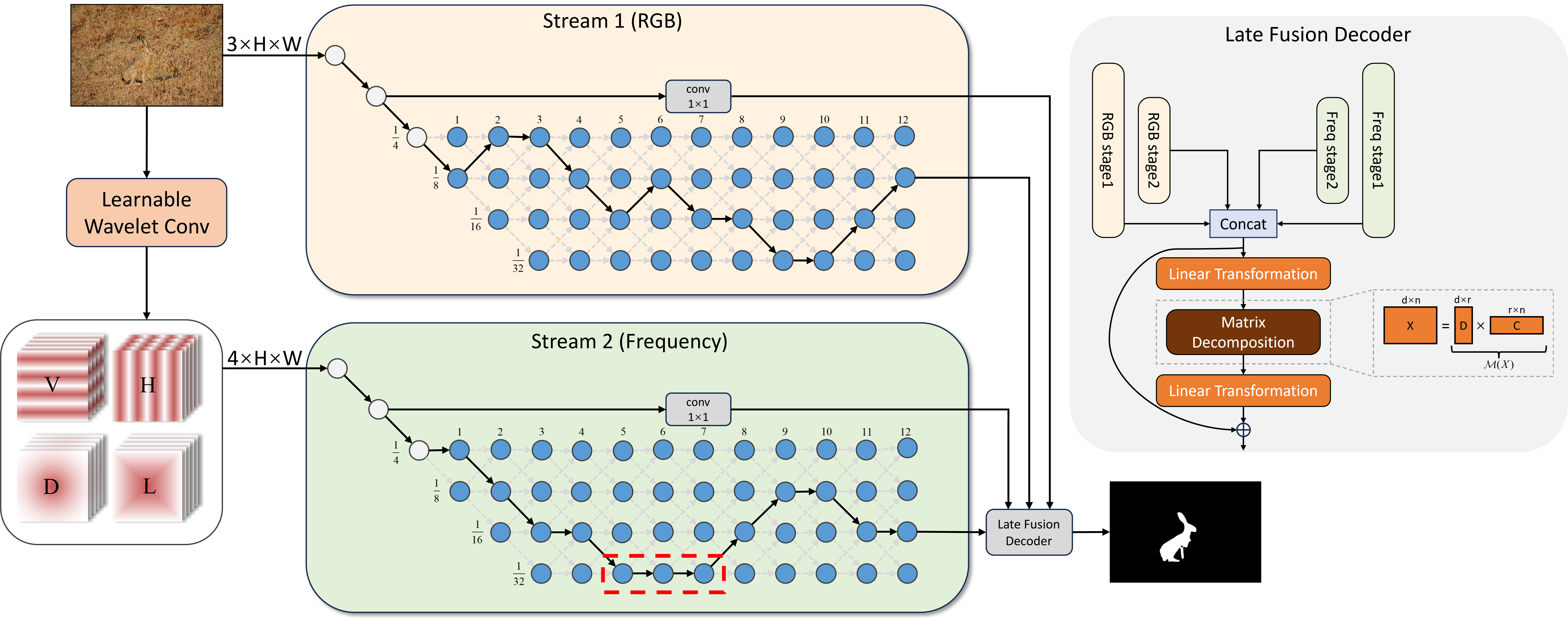}
	\end{center}
	\caption{Overview of CamoNAS. A learnable wavelet convolution (LDWT) splits the image into four sub-bands \((L,H,V,D)\), yielding a \(4\times H\times W\) frequency input that runs in parallel with RGB. The top and bottom backbones are two NAS-searched streams operating at four resolutions (\(\tfrac{1}{4}\), \(\tfrac{1}{8}\), \(\tfrac{1}{16}\), \(\tfrac{1}{32}\) of \(H\times W\)). Within each stage, blue nodes denote intermediate block states and edges denote candidate operations. At the network level, down/hold/up transitions between adjacent scales are jointly optimized with the cell operations in a single supernet, enabling dynamic multi-resolution paths. Selected multi-scale features from both streams are concatenated and fed to a late-fusion decoder to predict the mask, yielding sharper object boundaries and better detail preservation in camouflaged scenes.
	}
	\label{fig:overview}
\end{figure*}

\subsection{Architecture Overview}

As illustrated in Fig.~\ref{fig:overview}, CamoNAS comprises four components: (i) a learnable discrete wavelet transform (LDWT), (ii) two NAS-searched backbones (an RGB stream and a frequency stream), (iii) a network-level resolution path search, and (iv) a late-fusion head. The LDWT decomposes the input into one low-frequency map L and three directional high-frequency maps$\{H,V,D\}$, yielding a 4-channel tensor (\(4\times H\times W\)) that feeds the Frequency stream, while the original RGB image (\(3\times H\times W\)) feeds the RGB stream. Each stream is organized into four stages operating at spatial scales \(S=\{\tfrac{1}{4},\tfrac{1}{8},\tfrac{1}{16},\tfrac{1}{32}\}\) of \(H\times W\); within a stage, differentiable micro-cells update hidden states, where nodes denote block states and edges denote candidate operations. At the network level, down/hold/up transitions between adjacent scales are learned to form dynamic multi-resolution paths. Unlike hand-designed COD networks (e.g., SINet/DINet) that adopt a fixed, monotonic downsampling schedule with pre-defined fusion/refinement modules, CamoNAS learns the routing decisions (down/hold/up) from data and jointly optimizes them with cell structures. During training, we instantiate a supernet that contains all micro-cells and all scale transitions, and jointly optimize network weights and architecture parameters via continuous relaxation. After the final stage, selected features from both streams (e.g., \(F^{R}_{\text{stage}1}, F^{R}_{\text{stage}2}, F^{F}_{\text{stage}1}, F^{F}_{\text{stage}2}\)) are concatenated and passed through the \emph{Fusion Head}: a linear mapping, a low-rank matrix decomposition \(X\!\approx\!DC\), and a second projection that outputs a segmentation mask at the original image size. The entire model is trained with a structure-aware loss~\cite{qin2019basnet,wei2020f3net} to emphasize boundaries and fine details in camouflaged scenes.

\subsection{Learnable Wavelet Frequency Decomposition}
Existing frequency decomposition schemes rely on fixed filter templates~\cite{cong2023frequency,xie2023frequency,zhong2022detecting} (e.g., Laplacian pyramids, DFT). Their high/low frequency splits and directional responses remain static during training, making it difficult to adapt to varying texture amplitudes and spatial scales in camouflage scenarios. Moreover, such hard-coded filtering lacks reversibility constraints. In addition, recent COD studies have explored learnable wavelet-like decomposition~\cite{He2023Camouflaged}, demonstrating the effectiveness of frequency-aware representations. Motivated by these observations, we implement a learnable discrete wavelet transform as an end-to-end convolutional operator $\mathcal{W}(\,\cdot\,;\theta)$, which is updated in sync with the network's weights and architecture parameters during training.

We use two learnable 1D analysis filters $\mathbf{a}_0,\mathbf{a}_1 \in \mathbb{R}^{K}$ (each of length $K$).
From these, four separable 2D convolution kernels are formed by outer products:

\begin{equation}
	\begin{aligned}
		F_{ll} &= \mathbf{a}_0 \otimes \mathbf{a}_0^{\top}\quad
		F_{lh} &= \mathbf{a}_0 \otimes \mathbf{a}_1^{\top}\\
		F_{hl} &= \mathbf{a}_1 \otimes \mathbf{a}_0^{\top}\quad
		F_{hh} &= \mathbf{a}_1 \otimes \mathbf{a}_1^{\top}
	\end{aligned}
	\label{eq:filters}
\end{equation}
The transform performs the forward wavelet transform using a group convolution:
\begin{equation}
	X^{w}=\mathcal{W}(X;\theta)
	=\left\{X^{L},\,X^{H}_{h},\,X^{H}_{v},\,X^{H}_{d}\right\}
\end{equation}

The four groups of output features correspond to the low-frequency component, horizontal high-frequency, vertical high-frequency, and diagonal high-frequency components. We implement the dyadic downsampling with a stride-2 group convolution, thus no additional pooling is required. To ensure that the wavelet transform and its inverse form a stable invertible pair early in training, we tie the parameters of the analysis and synthesis filters and impose two perfect reconstruction constraints:
\begin{equation}
	\begin{aligned}
		A_{0}(z)S_{0}(z) + A_{1}(z)S_{1}(z) &= 2 \\
		A_{0}(-z)S_{0}(z) + A_{1}(-z)S_{1}(z) &= 0
	\end{aligned}
	\label{eq:yueshu}
\end{equation}
$A_{k}(z)$ and $S_{k}(z)$ are the $z^{-}$transforms of the corresponding filters, which are explicitly regularized in the frequency domain to ensure lossless round trips and suppress aliasing. We incorporate these constraints into the training objective as penalty terms. They add minimal overhead to the training process but significantly improve the network's ability to preserve high frequency information.

\subsection{Hierarchical and Multi-Scale Search}
\label{subsec:hierarchical}

\subsubsection{Cell-Level Search}
\label{subsubsec:cell}

\begin{figure*}
	\begin{center}
		\includegraphics[width=13cm]{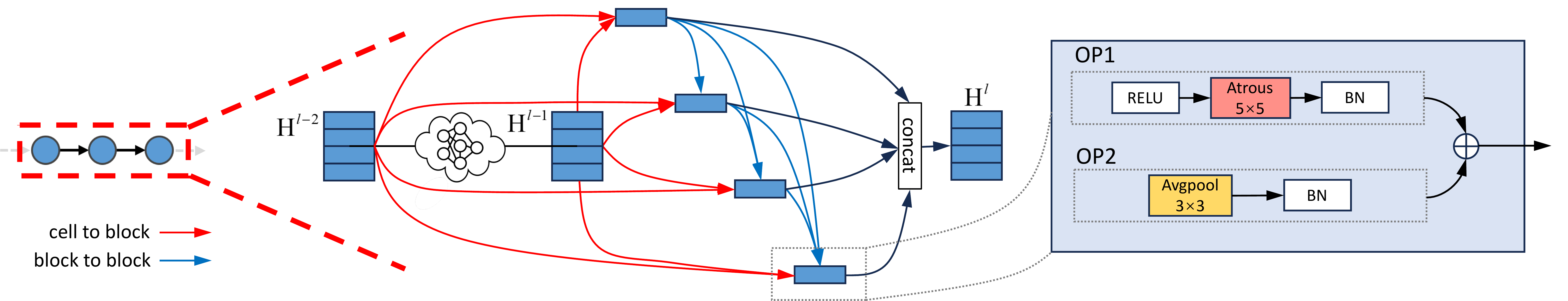}
	\end{center}
	\caption{Cell-level search space of CamoNAS. 
		In the $l$-th cell, the candidate input set consists of the two preceding cell outputs $(H^{l-2},\,H^{l-1})$ and all earlier block states within the cell (\textcolor{red}{red}: cell$\!\to$block, \textcolor{blue}{blue}: block$\!\to$block). 
		Each block selects two inputs $I_1,I_2\!\in\!C$, applies two operators $O_1,O_2\!\in\!\mathcal{O}$ (examples on the right), and outputs $H^i=O_1(I_1)+O_2(I_2)$; edges are weighted by architecture parameters $\alpha$ under continuous relaxation during search. 
		The outputs of all active blocks are concatenated to form the cell output $H^{l}$.}
	\label{fig:cell}
\end{figure*}

Similar to DARTS-style architectures~\cite{xu2019pc,chen2019progressive,chu2020darts,liu2018darts}, we construct the network as a hierarchy of multiple cells stacked sequentially. Each cell serves as a basic building block and contains several internal blocks. We model a cell as a directed acyclic graph (DAG) and adopt continuous relaxation to enable efficient gradient-based optimization of its discrete structure, facilitating the search over a large architecture space. Below, we detail the design of the cell-level search space. The overall cell-level search space is illustrated in Fig.~\ref{fig:cell}.

Suppose a cell contains $n$ blocks, and let $H^{i}$ denote the output of the $i$-th block (for $i = 1, 2, \dots, n$). In each block, we first select two input features from a candidate set $C$ (denote them as $I_{1}$ and $I_{2}$ as in the diagram). We then apply two operators $O_{1}$ and $O_{2}$ to these features respectively, and finally sum the two resulting tensors element-wise to obtain the block's output $H^{i}=O_{1}(I_{1})+O_{2}(I_{2})$.

Here, $O_{1}$ and $O_{2}$ are the chosen operations for inputs $I_{1}$ and $I_{2}$, respectively, and $I_{1}, I_{2} \in C$ are the two inputs selected from the candidate set $C$. The candidate input set $C$ is defined as: 
\begin{equation}
	C = \bigl\{ H_{\text{cell-}1},\, H_{\text{cell-}2} \bigr\}
	\;\cup\;
	\bigl\{ H^{1},\, H^{2},\, \dots,\, H^{i-1} \bigr\}
\end{equation}
Specifically, $H_{cell-1}$ and $H_{cell-2}$ represent the outputs from the previous and the preceding-to-previous cells, respectively. $\{H^1, ..., H^{i-1}\}$ denotes the outputs of all internal blocks within the current cell.

For the selected inputs $I_1$ and $I_2$, we separately apply operations $O_1$ and $O_2$ for feature transformations. Based on a comprehensive consideration of existing COD research~\cite{fan2021concealed}, we set our candidate operation set $O$ to include eight widely used convolutional operations: two depthwise separable convolutions, two atrous convolutions (with dilation $r=2$), average pooling, max pooling, skip connection, and no connection. These fixed operations enhance feature extraction capabilities from multiple perspectives while maintaining computational simplicity. Moreover, inspired by recent frequency-sensitive filters commonly accelerated via GPU~\cite{huang2023adaptive,he2025frequency}, we incorporate directional edge filters (Sobel), Haar wavelet split, and Gaussian blur to further enhance the model's capacity in capturing edge targets. Given that depthwise separable convolutions have been proven sufficient to serve as parameter-efficient benchmarks in recent NAS research, we no longer separately include ordinary convolution, thus maintaining conciseness of the search space. Additionally, by combining fixed operations with learnable components, we maintain simplicity while expanding the versatility of operations. Such a design ensures stable dimensionality and consistent gradient flow. The final output of each cell $H_{cell}$ is obtained by concatenating the outputs of all blocks along the channel dimension, as follows:
\begin{equation}
	H_{cell} = \operatorname{Concat}_{\mathrm{channel}}(H^1, H^2, ..., H^i)
\end{equation}

During the search process, we adopt a continuous relaxation strategy for the candidate operations, converting discrete operation choices into learnable continuous parameters. Specifically, for each candidate operation $o$ in operation set $O$, we introduce a parameter $\alpha_o$. After applying a softmax normalization, the output for the $j$-th input $I_j$ in the $i$-th block is computed as a weighted sum of all candidate operations:
\begin{equation}
	O_j(I_j) = \sum_{o\in O} \frac{\exp(\alpha_o)}{\sum_{o' \in O}\exp(\alpha_{o'})} \cdot o(I_j)
\end{equation}

Such continuous relaxation not only enables differentiable cell-level search, but also allows us to leverage gradient-based optimization to update the architecture parameters end-to-end, progressively exploring candidate architectures that fulfill task-specific requirements from coarse to fine levels.
\subsubsection{Layer-level Search}

Based on the constructed cell-level search space, we further design the overall network architecture consistent with the multi-resolution search strategy adopted in micro-architecture-level searches. The network consists of multiple stacked cells, dynamically adjusting the spatial resolution of feature maps between layers to meet the dual demands of fine-grained edge detection and global semantic comprehension. In our design, the initial network front end adopts two layers of convolution for preliminary feature extraction and downsampling, producing initial hierarchical features. Then, the network comprises $L$ cells, each layer $l \in \{1, \dots, L\}$ operating at four discrete resolutions:

\begin{equation}
	S = \left\{ s_1=\frac{1}{4}, s_2=\frac{1}{8}, s_3=\frac{1}{16}, s_4=\frac{1}{32} \right\}
\end{equation}
where each cell explicitly searches its resolution paths and cross-layer information flow.

To ensure continuity of spatial resolution across layers, we impose strict constraints on the resolution transitions between layers. Specifically, the state of layer $l$ can only be derived from states of the preceding layer $(l-1)$ through $\{\times 1, \times 2, \times \frac{1}{2}\}$ scaling operations:

\begin{equation}
	s^l \in \left\{ s^{l-1}, 2s^{l-1}, \frac{1}{2}s^{l-1} \right\} \cap S
\end{equation}
The highest resolution is fixed at $s_1=\frac{1}{4}$, and the lowest resolution is $s_4=\frac{1}{32}$.

To introduce differentiability into discrete paths, for layer $l$, we define a learnable parameter $\beta_{s \rightarrow s'}^{l}$ for transitions from scale $s$ to $s'$. Through softmax normalization, we obtain transition probabilities:

\begin{equation}
	\pi^{\,l}_{s\to s'} =
	\frac{\exp(\beta^{\,l}_{s\to s'})}{\sum_{s''\in N(s)} \exp(\beta^{\,l}_{s\to s''})}
\end{equation}
Where $N(s)=\{s,2s,\tfrac12 s\}\cap S$. During forward propagation, the hidden state at scale $s$ for layer $l-1$ is denoted as $\mathbf{H}_s^{l-1}$. It is mapped to the target scale $s'$ through upsampling or downsampling operators (bilinear interpolation), denoted as: $R(s \rightarrow s')$ and subsequently passed into the current cell:

\begin{equation}
	\tilde{\mathbf{H}}_{s'}^{l} = \sum_{s \in N(s')} \pi_{s \rightarrow s'}^{l} R(s \rightarrow s')(\mathbf{H}_s^{l-1})
\end{equation}
The inputs $\tilde{\mathbf{H}}_{s'}^{l}$ at various resolutions are collectively fed into the $l$-th cell, yielding new hidden states $\mathbf{H}_{s'}^{l}$. This iterative approach jointly optimizes resolution paths and cell-internal operation selection.

In summary, the layer-level search learns when to keep, upsample, or downsample, producing data-driven multi-resolution paths instead of a fixed schedule. This lets high-resolution features preserve fine boundaries while low-resolution features consolidate global semantics in a complementary way. The resulting multi-scale states feed directly into the RGB-frequency fusion, enabling accurate camouflaged-object segmentation with modest complexity.

\subsection{Fusion Head}

In the previous sections, we have separately exploited rich multi-scale features from RGB and frequency-domain streams via Cell-NAS units. However, effectively integrating these cross-domain and cross-layer features remains challenging because direct concatenation often introduces information redundancy and semantic inconsistency between high- and low-level features, which degrades precise segmentation of camouflaged objects~\cite{fan2021concealed}. Therefore, we adopt a lightweight decoding module, designed as a low-dimensional embedding-based decoder. The Fusion Head adaptively integrates features while preserving key details and suppressing noise simultaneously.

Specifically, we first resize all selected features to the same spatial resolution, align channels with $1{\times}1$ projections, and flatten spatial dimensions so that $n=H\times W$. We then define:
\begin{equation}
	X=\operatorname{Concat}\!\big(F^{R}_{\text{stage}1},\,F^{R}_{\text{stage}2},\,F^{F}_{\text{stage}1},\,F^{F}_{\text{stage}2}\big)
\end{equation}

To effectively merge these multi-stage features, the Fusion Head applies a linear projection $\Phi_1$ to obtain a compact embedding with a common channel size $d$:
\begin{equation}
	Z=\Phi_1(X)\in\mathbb{R}^{d\times n}
\end{equation}
The embedding is then factorized by a low-rank module:
\begin{equation}
	Z \approx D\,C \qquad D\in\mathbb{R}^{d\times r}\; C\in\mathbb{R}^{r\times n},\; r\ll d
\end{equation}

We employ a soft vector quantization (Soft-VQ)~\cite{geng2021attention} scheme to obtain a low-rank approximation. Soft-VQ first computes cosine scores between dictionary atoms and column features:
\begin{equation}
	s_{ij}=\frac{d_i^\top z_j}{\|d_i\|\,\|z_j\|} \qquad i=1,\dots,r,\;\; j=1,\dots,n
\end{equation}

Membership weights are obtained via a temperature-controlled softmax over atoms:
\begin{equation}
	C_{ij}=\frac{\exp(s_{ij}/T)}{\sum_{k=1}^{r}\exp(s_{kj}/T)}
\end{equation}

The compact embedding is then reconstructed as \(\hat{Z}=DC\).
A second linear projection \(\Phi_2\) restores channels and forms a residual with \(Z\):
\begin{equation}
	Y=\Phi_2(\hat{Z})+Z
\end{equation}

Finally, after reshaping \(Y\) back to feature maps, BatchNorm and a two-layer \(1{\times}1\) MLP produce
the segmentation mask \(M\); when the decoder operates at a reduced scale, we apply bilinear upsampling
to match the original image resolution.

\subsection{Search and Training Strategy}

The optimization procedure of CamoNAS is divided into two sequential stages~\cite{liu2019autodeeplab}. The first stage jointly optimizes the network topology (cell operators and multi-resolution routing) while learning the LDWT kernels end-to-end. The second stage retrains weights on the discretized final architecture. The entire process maintains end-to-end differentiability. Specifically, the internal operations within each cell are probabilistically controlled by \(\alpha\), cross-layer resolution transitions by \(\beta\), learnable wavelet kernels by \(\theta\), and ordinary convolution and normalization parameters by \(\omega\).

\noindent\textbf{Architecture Search Stage}:  In the joint search stage, continuous relaxation is adopted, where parameters \(\alpha\) and \(\beta\) after softmax normalization directly propagate forward; parameters \(\omega\) and \(\theta\) directly parameterize weights. We define the task loss as \(L_{seg}(\omega,\theta;\alpha,\beta)\), along with a wavelet completeness regularization:
\begin{equation}
	L_{\mathrm{wavelet}}(\theta)
	= \sum_{k=0}^{N-1}
	\Big(
	\big\langle a_{0},\, s_{0} \big\rangle_{k}
	+ \big\langle a_{1},\, s_{1} \big\rangle_{k}
	- \widehat{V}_{k}
	\Big)^{2}
\end{equation}

We then iteratively update parameters \((\omega,\theta)\) on the training subset \(D_{trainA}\) and update \((\alpha,\beta)\) on the validation subset \(D_{trainB}\):
\begin{equation}
	(\omega,\theta)\leftarrow (\omega,\theta)-\eta_{\omega}\nabla_{\omega,\theta}[L_{seg}+\lambda L_{wavelet}]
\end{equation}
\begin{equation}
	(\alpha,\beta)\leftarrow (\alpha,\beta)-\eta_{\alpha}\nabla_{\alpha,\beta}L_{seg}
\end{equation}
Here, \(\eta_{\omega}\) and \(\eta_{\alpha}\) use shared cosine annealing schedules. To avoid premature fluctuations, we freeze $\alpha,\beta$ for the first $T_0$ epochs and optimize only the network weights $(\omega,\theta)$.

\noindent\textbf{Discretization and Retraining Stage}: After completing the search, for each cell, the discretized architecture \(\hat{\alpha}\) is obtained by taking the argmax over softmax weights. For network-level discretization, the optimal path \(\hat{\beta}\) is solved via Viterbi decoding on the layer-scale resolution probability map. After discretization, parameters \(\theta^{*}\) are locked, and comprehensive retraining on the entire dataset is initiated. The total loss at this stage is:
\begin{equation}
	L_{total}(x,y)=L_{struct}(x,y)+\lambda L_{wavelet}(\theta^{*})
\end{equation}
Where \(L_{struct}\) maintains the structure-aware BCE+IoU form~\cite{qin2019basnet,wei2020f3net}, and \(L_{wavelet}\) merely serves as a constant regularizer for determined wavelet kernels, no longer impacting gradient updates. This two-stage strategy allows sufficient exploration of \(\alpha,\beta\) during the search stage and ensures that \(\theta\) does not degenerate after discretization. Consequently, the obtained network effectively exploits complementarity between RGB and frequency-domain streams, significantly improving segmentation accuracy on object boundaries and fine details.

\section{Experiments}
\label{sec:Experiments}
\subsection{Training Settings}

To ensure reproducibility and fairness, we strictly follow widely accepted NAS experimental setups~\cite{zhong2022detecting,He2023Camouflaged}, dividing training into architecture search and final-model retraining.

\noindent\textbf{Architecture Search Stage.} The training dataset is randomly split into two equally sized subsets \(D_{train}^{A}\) and \(D_{train}^{B}\), each containing 2020 images. Network weights \((\omega,\theta)\) are updated on \(D_{train}^{A}\). Architecture parameters \((\alpha,\beta)\) are updated using Adam on \(D_{train}^{B}\) with an initial learning rate of \(3\times10^{-4}\), for 60 epochs with batch size 2. Both architecture and weight optimizers use cosine-annealing schedules. We empirically observed that updating the architecture parameters \((\alpha,\beta)\) from the outset often traps the search in local optima. Accordingly, we adopt a warm-up schedule that optimizes only the network/LDWT weights \((\omega,\theta)\) for the first 20~epochs before enabling updates to \((\alpha,\beta)\); the full search used 4 GPU-days on two NVIDIA RTX 2080~Ti GPUs.

\noindent\textbf{Discretization and Retraining Stage.} After the search, we retrain the discretized architecture from scratch on the full training set for 90 epochs with SGD
(initial LR $1\times10^{-3}$; step decay $\times0.5$ every 20 epochs; momentum $0.9$; weight decay $1\times10^{-5}$; batch size 16).

All training images are resized and randomly cropped to $512\times512$, with random horizontal flipping, scale jitter (0.8-1.2$\times$), and color jitter for data augmentation~\cite{fan2020camouflaged,yang2021uncertainty}.
Our implementation is in PyTorch and runs on two NVIDIA RTX 2080~Ti GPUs.

\subsection{Datasets}
To rigorously evaluate the generalization and robustness of our model, we conduct experiments on four widely adopted benchmark datasets for COD, including \textit{CHAMELEON}, \textit{CAMO}, \textit{COD10K}, and \textit{NC4K}. Following the official splits used in prior works~\cite{pang2022zoom} to ensure an unbiased comparison, we train on 1,000 images from \textit{CAMO} and 3,040 images from \textit{COD10K}. The testing phase is conducted on the full testing subsets of the datasets mentioned above, ensuring comprehensive and balanced performance evaluation across diverse scenarios.

\subsection{Evaluation Metrics and Comparison}
\label{subsec:metrics}
We evaluate COD with four standard metrics:
structure measure ($S_{\alpha}$), mean absolute error (MAE, $M$), weighted F-measure ($F_{\beta}^\omega$) and mean E-measure ($E_{\phi}$).
We compare CamoNAS with over ten representative methods (e.g., SINet, JSCOD, PFNet, FDNet, SegMaR, ZoomNet, DGNet, DINet) on \textit{CHAMELEON}, \textit{CAMO}, \textit{COD10K}, and \textit{NC4K}.
For fairness, predictions for baselines are obtained from official releases or reproduced from their public code under the same protocols.

\begin{table*}[t]
	\centering
	\resizebox{\textwidth}{!}{%
		\setlength{\tabcolsep}{1mm}
		\begin{tabular}{l|c|cccc|cccc|cccc|cccc}
			\toprule
			\multirow{2}{*}{Methods} & \multirow{2}{*}{Pub.} &
			\multicolumn{4}{c|}{\textit{CHAMELEON}} &
			\multicolumn{4}{c|}{\textit{CAMO}} &
			\multicolumn{4}{c|}{\textit{COD10K}} &
			\multicolumn{4}{c}{\textit{NC4K}} \\
			\cline{3-18}
			
			& & 
			{\cellcolor{gray!40}$M\downarrow$} &
			{\cellcolor{gray!40}$F_{\beta}^{\omega}\uparrow$} &
			{\cellcolor{gray!40}$E_{\phi}\uparrow$} &
			{\cellcolor{gray!40}$S_{\alpha}\uparrow$} &
			{\cellcolor{gray!40}$M\downarrow$} &
			{\cellcolor{gray!40}$F_{\beta}^{\omega}\uparrow$} &
			{\cellcolor{gray!40}$E_{\phi}\uparrow$} &
			{\cellcolor{gray!40}$S_{\alpha}\uparrow$} &
			{\cellcolor{gray!40}$M\downarrow$} &
			{\cellcolor{gray!40}$F_{\beta}^{\omega}\uparrow$} &
			{\cellcolor{gray!40}$E_{\phi}\uparrow$} &
			{\cellcolor{gray!40}$S_{\alpha}\uparrow$} &
			{\cellcolor{gray!40}$M\downarrow$} &
			{\cellcolor{gray!40}$F_{\beta}^{\omega}\uparrow$} &
			{\cellcolor{gray!40}$E_{\phi}\uparrow$} &
			{\cellcolor{gray!40}$S_{\alpha}\uparrow$} \\
			\midrule
			SAM~\cite{kirillov2023segment}                                         & \multicolumn{1}{c|}{---}                              & 0.207                                 & 0.595                                 & 0.647                                 & 0.635                                 & 0.160                                 & 0.597                                 & 0.639                                 & 0.643                                 & 0.093                                 & 0.673                                 & 0.737                                 & 0.730                                 & 0.118                                 & 0.675                                 & 0.723                                 & 0.717                                 \\
			SAM-S~\cite{kirillov2023segment}                                         & \multicolumn{1}{c|}{---}                              & 0.076                                 & 0.729                                 & 0.820                                 & 0.650                                 & 0.105                                 & 0.682                                 & 0.774                                 & 0.731                                 & 0.046                                & 0.695                                 & 0.828                                 & 0.772                                 & 0.071                                 & 0.747                                 & 0.832                                 & 0.763                                 \\
			SINet~\cite{fan2020camouflaged}                                          & CVPR20                                             & 0.034                                 & 0.806                                 & 0.938                                 & 0.872                                 & 0.092                                 & 0.644                                 & 0.804                                 & 0.745                                 & 0.043                                 & 0.631                                 & 0.864                                 & 0.776                                 & 0.058                                 & 0.723                                 & 0.871                                 & 0.808                                 \\

			JSCOD~\cite{li2021uncertainty}                                          & CVPR21                                             & 0.030                                 & 0.848                                 & 0.943                                 & 0.894                                 & 0.073                                 & 0.728                                 & 0.859                                 & 0.800                                 & 0.035                                 & 0.684                                 & 0.884                                 & 0.809                                 & 0.047                                 & 0.771                                 & 0.898                                 & 0.842                                 \\

			PFNet~\cite{mei2021camouflaged}                                          & CVPR21                                             & 0.033                                 & 0.810                                 & 0.921                                 & 0.882                                 & 0.085                                 & 0.695                                 & 0.841                                 & 0.782                                 & 0.040                                 & 0.660                                 & 0.877                                 & 0.800                                 & 0.053                                 & 0.745                                 & 0.887                                 & 0.829                                 \\

			FDNet~\cite{zhong2022detecting}                                          & CVPR22                                             & 0.030                                 & 0.819                                 & 0.948                                 & 0.894                                 & 0.063                                & 0.775                                 & 0.895                                 & \textcolor{mygreen}{\textbf{0.841}}                                 & 0.030                                 & 0.729                                 & 0.919                                 & 0.840                                 & 0.052                                 & 0.750                                 & 0.893                                 & 0.834                                 \\

			SegMaR~\cite{jia2022segment}                                          & CVPR22                                             & 0.027                                 & 0.835                                 & 0.950                                 & 0.897                                 & 0.071                                 & 0.753                                 & 0.874                                 & 0.815                                 & 0.034                                 & 0.724                                 & 0.899                                 & 0.833                                 & 0.046                                 & 0.781                                 & 0.896                                 & 0.841                                 \\

			ZoomNet~\cite{pang2022zoom}                                          & CVPR22                                             & 0.023                                & 0.845                                 & 0.943                                 & 0.902                                 & 0.066                                 & 0.752                                 & 0.877                                 & 0.820                                 & 0.029                                 & 0.729                                 & 0.888                                 & 0.838                                 & 0.043                                 & 0.784                                 & 0.896                                 & 0.853                                 \\

			DGNet~\cite{ji2023deep}                                          & MIR23                                             & 0.029                                 & 0.816                                 & 0.934                                 & 0.890                                 & \textcolor{myred}{\textbf{0.057}}                                 & 0.769                                 & \textcolor{mygreen}{\textbf{0.901}}                                 & \textcolor{myblue}{\textbf{0.839}}                                 & 0.033                                 & 0.693                                 & 0.896                                 & 0.822                                 & 0.042                                 & 0.784                                 & 0.911                                 & 0.857                                 \\

			PopNet~\cite{wu2023source} & ICCV23
			& {\textcolor{mygreen}{\textbf{0.022}}}
			& {\textcolor{mygreen}{\textbf{0.893}}}
			& {\textcolor{myblue}{\textbf{0.962}}}
			& {\textcolor{myblue}{\textbf{0.910}}}
			& 0.077
			& 0.744
			& 0.859
			& 0.808
			& {\textcolor{myblue}{\textbf{0.028}}}
			& 0.757
			& 0.910
			& \textcolor{myblue}{\textbf{0.851}}
			& 0.042
			& 0.802
			& 0.910
			& {\textcolor{mygreen}{\textbf{0.861}}} \\

			FEDER~\cite{He2023Camouflaged}                                          & CVPR23                                             & 0.028                                 & 0.855                                 & 0.947                                 & 0.894                                 & 0.069                                 & \textcolor{myblue}{\textbf{0.785}}                                 & 0.873                                 & 0.807                                 & 0.032                                 & 0.740                                 & 0.900                                 & 0.823                                 & 0.045                                 & \textcolor{myblue}{\textbf{0.817}}                                 & 0.905                                 & 0.846                                 \\
			
			SCOD~\cite{he2022weakly}                                         & AAAI23                                             & 0.046 & 0.791 & 0.897 & 0.818                                 & 0.092 & 0.709                                 & 0.815                                 & 0.735                                 & 0.049                                 & 0.637                                 & 0.832                                 & 0.733                                 & 0.064                                 & 0.751                                 & 0.853                                 & 0.779                                 \\

			WS-SAM~\cite{he2023weakly}                                          & NeurIPS24                                             & 0.046                                 & 0.777                                 & 0.897                                 & 0.824                                 & 0.090                                 & 0.716                                 & 0.818                                 & 0.759                                 & 0.038                                 & 0.719                                 & 0.878                                 & 0.803                                 & 0.052                                 & 0.802                                 & 0.886                                 & 0.829                                 \\

			ICEG~\cite{he2023strategic}                                          & ICLR24                                             & 0.027                                 & 0.858                                 & 0.950                                 & 0.899                                 & 0.068                                 & \textcolor{mygreen}{\textbf{0.789}}                                 & 0.879                                 & 0.810                                 & 0.030                                 & 0.747                                 & 0.906                                 & 0.826                                 & 0.044                                 & 0.814                                 & 0.908                                 & 0.849                                 \\

			DINet~\cite{zhou2024decoupling}                                          & TMM24                                            & -                                 & -                                 & -                                 & -                                 & \textcolor{myblue}{\textbf{0.063}}                                 & 0.775                                 & \textcolor{myblue}{\textbf{0.895}}                                 & \textcolor{myred}{\textbf{0.841}}                                 & 0.030                                 & 0.729                                 & \textcolor{myblue}{\textbf{0.919}}                                 & 0.840                                 & 0.052                                 & 0.750                                 & 0.893                                 & 0.834                                 \\

			ZoomNeXt~\cite{pang2024zoomnext}                                          & TPAMI24                                             & {\textcolor{myred}{\textbf{0.020}}}                                 & \textcolor{myblue}{\textbf{0.863}}                                 & \textcolor{myred}{\textbf{0.969}}                                 & \textcolor{myred}{\textbf{0.912}}                                 & 0.069                                 & 0.760                                 & 0.885                                 & 0.822                                 & \textcolor{myred}{\textbf{0.026}}                                 & \textcolor{myblue}{\textbf{0.758}}                                 & \textcolor{mygreen}{\textbf{0.926}}                                 & \textcolor{myred}{\textbf{0.855}}                                 & \textcolor{myred}{\textbf{0.038}}                                 & 0.808                                 & \textcolor{myred}{\textbf{0.925}}                                 & \textcolor{myred}{\textbf{0.869}}                                 \\

			GenSAM~\cite{hu2024relax}                                          & AAAI24                                             & 0.090                                 & 0.680                                 & 0.807                                 & 0.764                                 & 0.113                                 & 0.659                                 & 0.775                                 & 0.719                                 & 0.067                                 & 0.681                                 & 0.838                                 & 0.775                                 & -                                 & -                                 & -                                 & -                                 \\

			DINO~\cite{yan2025ucod}                                          & CVPR25                                             & 0.031                                 & 0.825                                 & 0.931                                 & 0.864                                 & 0.077                                 & 0.747                                 & 0.862                                 & 0.793                                 & 0.031                                 & \textcolor{mygreen}{\textbf{0.763}}                                 & 0.916                                 & 0.834                                 & 0.043                                 & \textcolor{mygreen}{\textbf{0.818}}                                 & \textcolor{mygreen}{\textbf{0.923}}                                 & 0.850                                 \\

			CIRCOD~\cite{gupta2025circod}                                          & WACV25                                             & -                                 & -                                 & -                                 & -                                 & 0.063                                 & 0.772                                 & 0.894                                 & 0.824                                 & 0.030                                 & 0.741                                 & 0.916                                 & 0.835                                 & \textcolor{mygreen}{\textbf{0.040}}                                 & 0.808                                 & \textcolor{myblue}{\textbf{0.920}}                               & 0.858                                \\

			\rowcolor{c2!20}Ours & \multicolumn{1}{c|}{---} & {\textcolor{myblue}{\textbf{0.023}}} & \textcolor{myred}{\textbf{0.904}}    & \textcolor{mygreen}{\textbf{0.966}} & \textcolor{mygreen}{\textbf{0.911}} & \textcolor{mygreen}{\textbf{0.062}} & \textcolor{myred}{\textbf{0.808}} & \textcolor{myred}{\textbf{0.902}} & 0.831 & \textcolor{mygreen}{\textbf{0.027}} & \textcolor{myred}{\textbf{0.798}} & \textcolor{myred}{\textbf{0.926}} & \textcolor{mygreen}{\textbf{0.853}} & \textcolor{myblue}{\textbf{0.041}} & \textcolor{myred}{\textbf{0.827}} & 0.918 & \textcolor{myblue}{\textbf{0.858}} \\ \midrule
		\end{tabular}
	}%
	\caption{Results on COD. The best three results are highlighted in
		\textbf{\textcolor{myred}{red}},
		\textbf{\textcolor{mygreen}{green}} and
		\textbf{\textcolor{myblue}{blue}}.}
	\label{table:CODQuanti}
\end{table*}

\begin{figure}  
	\centering
	\includegraphics[width=\linewidth]{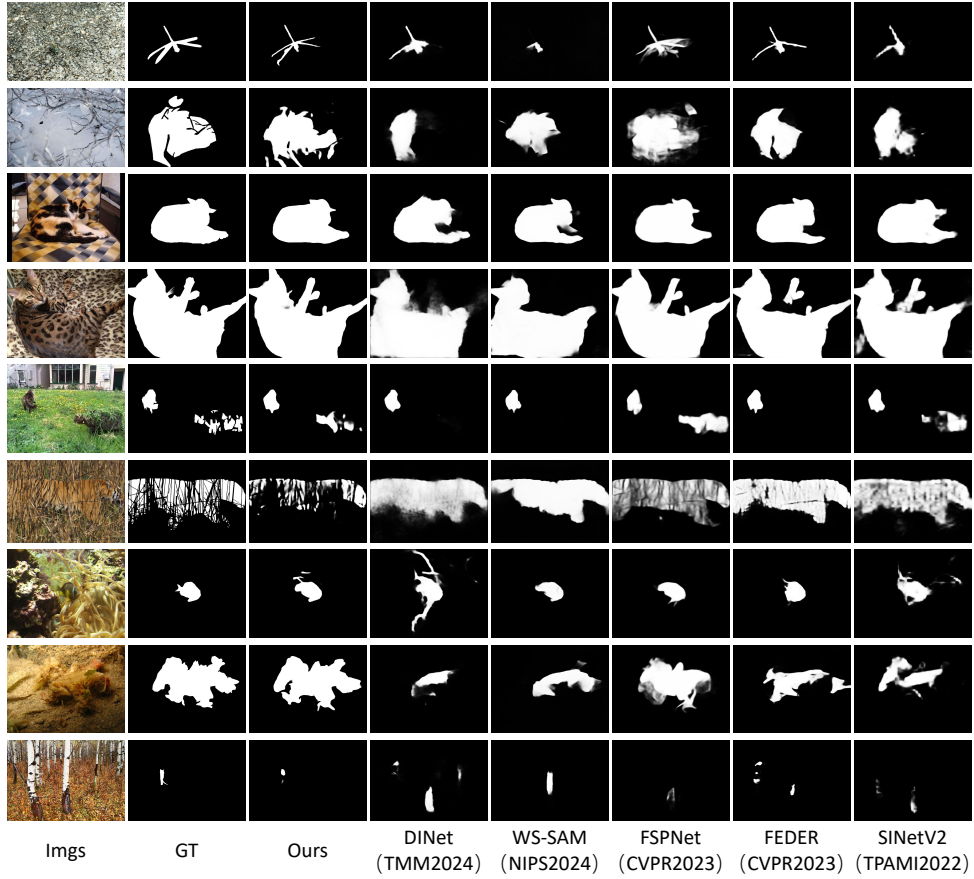}
	\caption{Qualitative comparisons of our CamoNAS with state-of-the-art methods.}
	\label{fig:Qualitative}
\end{figure}

\subsection{Experimental Results}

\textbf{Quantitative.} Table~\ref{table:CODQuanti} summarizes results on \textit{CHAMELEON}, \textit{CAMO}, \textit{COD10K}, and \textit{NC4K}. 
CamoNAS surpasses strong baselines (e.g., DINet, PopNet) on most metrics and datasets, delivering strong overall performance across $M$, $F_{\beta}^\omega$, $E_\phi$, and $S_\alpha$, with especially notable gains on \textit{CHAMELEON} and \textit{COD10K}.

\noindent\textbf{Qualitative.} Fig.~\ref{fig:Qualitative} shows representative cases from \textit{CHAMELEON}, \textit{CAMO}, \textit{COD10K}, and \textit{NC4K}. The panel spans typical COD challenges small or huge targets, multiple instances, occlusion, and boundary uncertainty. CamoNAS recovers finer structures and boundaries while better suppressing cluttered backgrounds, highlighting robustness under diverse camouflage scenarios.

\begin{table*}[t]
	\centering
	\begin{subtable}[t]{0.448\textwidth}
		\centering
		\resizebox{\columnwidth}{!}{
			\setlength{\tabcolsep}{1mm}
			\begin{tabular}{cccc|cccc}
				\toprule
				RGB & Freq & LDWT & $L_{\text{wav}}$ & $M$~$\downarrow$ & $F_{\beta}^\omega$~$\uparrow$ & $E_\phi$~$\uparrow$ & $S_\alpha$~$\uparrow$ \\ \midrule
				\checkmark  &  &   &  & 0.083 & 0.552 & 0.748 & 0.690 \\
				\checkmark & \checkmark  &   &  & 0.052 & 0.641 & 0.834 & 0.753 \\
				\checkmark &\checkmark &\checkmark  & & 0.039 & 0.726 & 0.887 & 0.805 \\
				\rowcolor{c2!20}\checkmark & \checkmark&\checkmark  &\checkmark & \textbf{0.027} & \textbf{0.798} & \textbf{0.926} & \textbf{0.853} \\ \bottomrule
		\end{tabular}}
		\caption{Dual-Stream Frequency Modeling}
		\label{tab:abl_a}
	\end{subtable}
	\begin{subtable}[t]{0.512\textwidth}
		\centering
		\resizebox{\columnwidth}{!}{
			\setlength{\tabcolsep}{1mm}
			\begin{tabular}{cccc|cccc}
				\toprule
				$\mathcal{O}_{\text{base}}$ & Sobel & HaarSplit & Gaussian & $M$~$\downarrow$ & $F_{\beta}^\omega$~$\uparrow$ & $E_\phi$~$\uparrow$ & $S_\alpha$~$\uparrow$ \\ \midrule
				\checkmark  &  &   &  & 0.037 & 0.755 & 0.892 & 0.824 \\
				\checkmark & \checkmark  &   &  & 0.033 & 0.752 & 0.904 & 0.822 \\
				\checkmark &\checkmark &\checkmark  & & 0.030 & 0.776 & 0.913 & 0.840 \\
				\rowcolor{c2!20}\checkmark & \checkmark&\checkmark  &\checkmark & \textbf{0.027} & \textbf{0.798} & \textbf{0.926} & \textbf{0.853} \\ \bottomrule
		\end{tabular}}
		\caption{Operation Space $\mathcal O$}
		\label{tab:abl_b}
	\end{subtable}
	
	\begin{subtable}[t]{0.425\textwidth}
		\centering
		\resizebox{\columnwidth}{!}{
			\setlength{\tabcolsep}{1mm}
			\begin{tabular}{c|ccccc}
				\toprule
				Metrics & Fixed $1/8$ & Fixed $1/16$ & Fixed 4-scale & Greedy & \cellcolor{c2!20}Ours   \\ \midrule
				$M$~$\downarrow$  & 0.107 & 0.118 & 0.041 & 0.052 & \cellcolor{c2!20}\textbf{0.027} \\
				$F_{\beta}^\omega$~$\uparrow$ & 0.368 & 0.247  & 0.713 & 0.625  & \cellcolor{c2!20}\textbf{0.798} \\
				$E_\phi$~$\uparrow$ & 0.678 & 0.669 & 0.872 & 0.831  & \cellcolor{c2!20}\textbf{0.926} \\
				$S_\alpha$~$\uparrow$ & 0.644 & 0.621 & 0.798 & 0.743  & \cellcolor{c2!20}\textbf{0.853} \\ \bottomrule
		\end{tabular}}
		\caption{Cross-Scale Path Search}
		\label{tab:abl_c}
	\end{subtable}
	\begin{subtable}[t]{0.535\textwidth}
		\centering
		\resizebox{\columnwidth}{!}{
			\setlength{\tabcolsep}{1mm}
			\begin{tabular}{c|cccc}
				\toprule
				Metrics & Concat+1$\times$1  & Hamburger~\cite{geng2021attention}   & Soft-VQ (r=16) & \cellcolor{c2!20}Soft-VQ (r=32, ours) \\ \midrule
				$M$~$\downarrow$    & 0.049 & 0.039 & 0.031 & \cellcolor{c2!20}\textbf{0.027} \\
				$F_{\beta}^\omega$~$\uparrow$ & 0.665 & 0.708 & 0.774 & \cellcolor{c2!20}\textbf{0.798} \\
				$E_\phi$~$\uparrow$ & 0.847 & 0.878 & 0.915 & \cellcolor{c2!20}\textbf{0.926} \\
				$S_\alpha$~$\uparrow$ & 0.762 & 0.795 & 0.835 & \cellcolor{c2!20}\textbf{0.853} \\ \bottomrule
		\end{tabular}}
		\caption{Fusion Head}
		\label{tab:abl_d}
	\end{subtable}
	
	\caption{Ablation studies of (a) Dual-Stream Frequency Modeling, (b) Operation Space $\mathcal O$, (c) Cross-Scale Path Search, (d) Fusion Head.}
	
	\label{tab:abl_all}
\end{table*}

\subsection{Ablation Study and Analysis}
\label{sec:ablation}

For a fair comparison, all variants are trained and evaluated under the same codebase, schedules, and augmentations. 
Unless otherwise stated, we train on \textit{CAMO} (1,000 images) and \textit{COD10K} (3,040 images), and report results on COD10K-Test (2026 images). 
All runs are conducted on two NVIDIA RTX 2080~Ti GPUs with identical hyper-parameters. 
We ablate one factor at a time and report $M$, $F_{\beta}^\omega$, $E_\phi$, and $S_\alpha$.

\noindent\textbf{Dual-stream frequency modeling.}
Adding a frequency branch to the RGB stream consistently improves performance across benchmarks (Table~\ref{tab:abl_a}). 
LDWT stabilizes sub-band decomposition, and a mild wavelet regularizer $L_{\text{wav}}$ further helps by preserving informative high/low-frequency structure.

\noindent\textbf{Operator space $\mathcal{O}$.}
Expanding $\mathcal{O}_{\text{base}}$ with Sobel, HaarSplit, and Gaussian yields consistent but modest gains (Table~\ref{tab:abl_b}), serving as a generic low-level operator vocabulary.
Notably, NAS still determines whether and where to use these primitives across edges/stages/scales.
Compared with operator enrichment (Table~\ref{tab:abl_b}), replacing the searched cross-scale routing leads to a much larger drop (Table~\ref{tab:abl_c}), indicating that the main improvement comes from routing/topology discovery rather than the operators alone.

\noindent\textbf{Cross-scale path search.}
Fixed single-scale or fixed four-scale schedules underuse cross-scale complementarity, and greedy routing is often locally optimal (Table~\ref{tab:abl_c}). 
Jointly searching down/hold/up transitions attains stronger structure-aware alignment while keeping MAE competitive. 
This controlled comparison indicates that the improvement is primarily driven by \emph{NAS-discovered routing structure and operator placement}, rather than merely training or hyperparameter tuning.

\noindent\textbf{Fusion head design.}
Concat+$1{\times}1$ leaves cross-stream redundancy and Hamburger~\cite{geng2021attention} adapts poorly to camouflage (Table~\ref{tab:abl_d}). 
Our low-rank \emph{Soft-VQ} decoder (best at $r{=}32$) improves all metrics via compact embeddings and soft quantization that preserve fine details.

\section{Conclusion}
\label{sec:conclusion}
We presented CamoNAS, a NAS framework for camouflaged object detection that jointly searches cell-level operators and network-level multi-resolution routing while pairing an RGB stream with a learnable wavelet-based frequency stream and a lightweight low-rank fusion head. This task-aligned search yields compact topologies that sharpen boundaries, suppress background distractions, and generalize across diverse camouflage patterns. Empirically, CamoNAS shows strong performance on four COD benchmarks, consistently outperforming recent methods in our evaluations. These results demonstrate that automated, COD-aware architecture design is a principled alternative to hand-designed pipelines. Future work will enlarge the search space to enable \emph{searchable} mid-level RGB--frequency coupling at intermediate stages (e.g., $1/8$ and $1/16$),
so that the two streams can interact during feature formation rather than only at the final fusion head, and we will explore transfer to related dense prediction tasks.


\bibliography{sn-bibliography}

@String(CVPR= {IEEE Conf. Comput. Vis. Pattern Recog.})

@String(ICCV= {Int. Conf. Comput. Vis.})

@String(ECCV= {Eur. Conf. Comput. Vis.})

@String(NIPS= {Adv. Neural Inform. Process. Syst.})

@String(ICPR = {Int. Conf. Pattern Recog.})

@String(ICASSP=	{ICASSP})

@String(AAAI = {AAAI})

@String(CVPR  = {CVPR})

@String(ICCV  = {ICCV})

@String(ECCV  = {ECCV})

@String(NIPS  = {NeurIPS})

@String(ICPR  = {ICPR})

@article{liu2018darts,
	title={Darts: Differentiable architecture search},
	author={Liu, Hanxiao and Simonyan, Karen and Yang, Yiming},
	journal={arXiv preprint arXiv:1806.09055},
	year={2018}
}

@inproceedings{liu2019autodeeplab,
	title={Auto-deeplab: Hierarchical neural architecture search for semantic image segmentation},
	author={Liu, Chenxi and Chen, Liang-Chieh and Schroff, Florian and Adam, Hartwig and Hua, Wei and Yuille, Alan L and Fei-Fei, Li},
	booktitle={Proceedings of the IEEE/CVF conference on computer vision and pattern recognition},
	pages={82--92},
	year={2019}
}

@article{bi2021rethinking,
	title={Rethinking camouflaged object detection: Models and datasets},
	author={Bi, Hongbo and Zhang, Cong and Wang, Kang and Tong, Jinghui and Zheng, Feng},
	journal={IEEE transactions on circuits and systems for video technology},
	volume={32},
	number={9},
	pages={5708--5724},
	year={2021},
	publisher={IEEE}
}

@article{mondal2017partially,
	title={Partially camouflaged object tracking using modified probabilistic neural network and fuzzy energy based active contour},
	author={Mondal, Ajoy and Ghosh, Susmita and Ghosh, Ashish},
	journal={International Journal of Computer Vision},
	volume={122},
	number={1},
	pages={116--148},
	year={2017},
	publisher={Springer}
}

@article{li2018fusion,
	title={A fusion framework for camouflaged moving foreground detection in the wavelet domain},
	author={Li, Shuai and Florencio, Dinei and Li, Wanqing and Zhao, Yaqin and Cook, Chris},
	journal={IEEE Transactions on Image Processing},
	volume={27},
	number={8},
	pages={3918--3930},
	year={2018},
	publisher={IEEE}
}

@article{le2019anabranch,
	title={Anabranch network for camouflaged object segmentation},
	author={Le, Trung-Nghia and Nguyen, Tam V and Nie, Zhongliang and Tran, Minh-Triet and Sugimoto, Akihiro},
	journal={Computer vision and image understanding},
	volume={184},
	pages={45--56},
	year={2019},
	publisher={Elsevier}
}

@article{fan2021concealed,
	title={Concealed object detection},
	author={Fan, Deng-Ping and Ji, Ge-Peng and Cheng, Ming-Ming and Shao, Ling},
	journal={IEEE transactions on pattern analysis and machine intelligence},
	volume={44},
	number={10},
	pages={6024--6042},
	year={2021},
	publisher={IEEE}
}

@inproceedings{mei2021camouflaged,
	title={Camouflaged object segmentation with distraction mining},
	author={Mei, Haiyang and Ji, Ge-Peng and Wei, Ziqi and Yang, Xin and Wei, Xiaopeng and Fan, Deng-Ping},
	booktitle={Proceedings of the IEEE/CVF conference on computer vision and pattern recognition},
	pages={8772--8781},
	year={2021}
}

@article{sun2022boundary,
	title={Boundary-guided camouflaged object detection},
	author={Sun, Yujia and Wang, Shuo and Chen, Chenglizhao and Xiang, Tian-Zhu},
	journal={arXiv preprint arXiv:2207.00794},
	year={2022}
}

@inproceedings{liu2022boosting,
	title={Boosting camouflaged object detection with dual-task interactive transformer},
	author={Liu, Zhengyi and Zhang, Zhili and Tan, Yacheng and Wu, Wei},
	booktitle={2022 26th International Conference on Pattern Recognition (ICPR)},
	pages={140--146},
	year={2022},
	organization={IEEE}
}

@inproceedings{pang2022zoom,
	title={Zoom in and out: A mixed-scale triplet network for camouflaged object detection},
	author={Pang, Youwei and Zhao, Xiaoqi and Xiang, Tian-Zhu and Zhang, Lihe and Lu, Huchuan},
	booktitle={Proceedings of the IEEE/CVF Conference on computer vision and pattern recognition},
	pages={2160--2170},
	year={2022}
}

@article{yin2024camoformer,
	title={Camoformer: Masked separable attention for camouflaged object detection},
	author={Yin, Bowen and Zhang, Xuying and Fan, Deng-Ping and Jiao, Shaohui and Cheng, Ming-Ming and Van Gool, Luc and Hou, Qibin},
	journal={IEEE Transactions on Pattern Analysis and Machine Intelligence},
	year={2024},
	publisher={IEEE}
}

@article{ji2023deep,
	title={Deep gradient learning for efficient camouflaged object detection},
	author={Ji, Ge-Peng and Fan, Deng-Ping and Chou, Yu-Cheng and Dai, Dengxin and Liniger, Alexander and Van Gool, Luc},
	journal={Machine Intelligence Research},
	volume={20},
	number={1},
	pages={92--108},
	year={2023},
	publisher={Springer}
}

@inproceedings{wu2023source,
	title={Source-free depth for object pop-out},
	author={Wu, Zongwei and Paudel, Danda Pani and Fan, Deng-Ping and Wang, Jingjing and Wang, Shuo and Demonceaux, C{\'e}dric and Timofte, Radu and Van Gool, Luc},
	booktitle={Proceedings of the IEEE/CVF international conference on computer vision},
	pages={1032--1042},
	year={2023}
}

@article{liang2024systematic,
	title={A systematic review of image-level camouflaged object detection with deep learning},
	author={Liang, Yanhua and Qin, Guihe and Sun, Minghui and Wang, Xinchao and Yan, Jie and Zhang, Zhonghan},
	journal={Neurocomputing},
	volume={566},
	pages={127050},
	year={2024},
	publisher={Elsevier}
}

@inproceedings{cong2023frequency,
	title={Frequency perception network for camouflaged object detection},
	author={Cong, Runmin and Sun, Mengyao and Zhang, Sanyi and Zhou, Xiaofei and Zhang, Wei and Zhao, Yao},
	booktitle={Proceedings of the 31st ACM international conference on multimedia},
	pages={1179--1189},
	year={2023}
}

@article{zhou2014low,
	title={Low-rank modeling and its applications in image analysis},
	author={Zhou, Xiaowei and Yang, Can and Zhao, Hongyu and Yu, Weichuan},
	journal={ACM Computing Surveys (CSUR)},
	volume={47},
	number={2},
	pages={1--33},
	year={2014},
	publisher={ACM New York, NY, USA}
}

@inproceedings{xie2023frequency,
	title={Frequency representation integration for camouflaged object detection},
	author={Xie, Chenxi and Xia, Changqun and Yu, Tianshu and Li, Jia},
	booktitle={Proceedings of the 31st ACM international conference on multimedia},
	pages={1789--1797},
	year={2023}
}

@inproceedings{zhai2021mutual,
	title={Mutual graph learning for camouflaged object detection},
	author={Zhai, Qiang and Li, Xin and Yang, Fan and Chen, Chenglizhao and Cheng, Hong and Fan, Deng-Ping},
	booktitle={CVPR},
	pages={12997--13007},
	year={2021}
}

@inproceedings{fan2020camouflaged,
	title={Camouflaged object detection},
	author={Fan, Deng-Ping and Ji, Ge-Peng and Sun, Guolei and Cheng, Ming-Ming and Shen, Jianbing and Shao, Ling},
	booktitle={CVPR},
	pages={2777--2787},
	year={2020}
}

@inproceedings{lv2021simultaneously,
	title={Simultaneously localize, segment and rank the camouflaged objects},
	author={Lv, Yunqiu and Zhang, Jing and Dai, Yuchao and Li, Aixuan and Liu, Bowen and Barnes, Nick and Fan, Deng-Ping},
	booktitle={CVPR},
	pages={11591--11601},
	year={2021}
}

@inproceedings{yang2021uncertainty,
	title={Uncertainty-guided transformer reasoning for camouflaged object detection},
	author={Yang, Fan and Zhai, Qiang and Li, Xin and Huang, Rui and Luo, Ao and Cheng, Hong and Fan, Deng-Ping},
	booktitle={ICCV},
	pages={4146--4155},
	year={2021}
}

@inproceedings{jia2022segment,
	title={Segment, Magnify and Reiterate: Detecting Camouflaged Objects the Hard Way},
	author={Jia, Qi and Yao, Shuilian and Liu, Yu and Fan, Xin and Liu, Risheng and Luo, Zhongxuan},
	booktitle={CVPR},
	pages={4713--4722},
	year={2022}
}

@inproceedings{ronneberger2015u,
	title={U-net: Convolutional networks for biomedical image segmentation},
	author={Ronneberger, Olaf and Fischer, Philipp and Brox, Thomas},
	booktitle={International Conference on Medical image computing and computer-assisted intervention},
	pages={234--241},
	year={2015},
	organization={Springer}
}

@article{chen2017deeplab,
	title={Deeplab: Semantic image segmentation with deep convolutional nets, atrous convolution, and fully connected crfs},
	author={Chen, Liang-Chieh and Papandreou, George and Kokkinos, Iasonas and Murphy, Kevin and Yuille, Alan L},
	journal={IEEE transactions on pattern analysis and machine intelligence},
	volume={40},
	number={4},
	pages={834--848},
	year={2017},
	publisher={IEEE}
}

@inproceedings{kirillov2020pointrend,
	title={Pointrend: Image segmentation as rendering},
	author={Kirillov, Alexander and Wu, Yuxin and He, Kaiming and Girshick, Ross},
	booktitle={Proceedings of the IEEE/CVF conference on computer vision and pattern recognition},
	pages={9799--9808},
	year={2020}
}

@inproceedings{lin2017feature,
	title={Feature pyramid networks for object detection},
	author={Lin, Tsung-Yi and Doll{\'a}r, Piotr and Girshick, Ross and He, Kaiming and Hariharan, Bharath and Belongie, Serge},
	booktitle={Proceedings of the IEEE conference on computer vision and pattern recognition},
	pages={2117--2125},
	year={2017}
}

@inproceedings{zhang2022preynet,
	title={Preynet: Preying on camouflaged objects},
	author={Zhang, Miao and Xu, Shuang and Piao, Yongri and Shi, Dongxiang and Lin, Shusen and Lu, Huchuan},
	booktitle={ACM MM},
	pages={5323--5332},
	year={2022}
}

@article{he2023weakly,
	title={Weakly-supervised concealed object segmentation with sam-based pseudo labeling and multi-scale feature grouping},
	author={He, Chunming and Li, Kai and Zhang, Yachao and Xu, Guoxia and Tang, Longxiang and Zhang, Yulun and Guo, Zhenhua and Li, Xiu},
	journal={Advances in Neural Information Processing Systems},
	volume={36},
	pages={30726--30737},
	year={2023}
}

@article{he2023strategic,
	title={Strategic preys make acute predators: Enhancing camouflaged object detectors by generating camouflaged objects},
	author={He, Chunming and Li, Kai and Zhang, Yachao and Zhang, Yulun and Guo, Zhenhua and Li, Xiu and Danelljan, Martin and Yu, Fisher},
	journal={arXiv preprint arXiv:2308.03166},
	year={2023}
}

@article{zhou2024decoupling,
	title={Decoupling and integration network for camouflaged object detection},
	author={Zhou, Xiaofei and Wu, Zhicong and Cong, Runmin},
	journal={IEEE Transactions on Multimedia},
	volume={26},
	pages={7114--7129},
	year={2024},
	publisher={IEEE}
}

@inproceedings{hu2024relax,
	title={Relax image-specific prompt requirement in sam: A single generic prompt for segmenting camouflaged objects},
	author={Hu, Jian and Lin, Jiayi and Gong, Shaogang and Cai, Weitong},
	booktitle={Proceedings of the AAAI Conference on Artificial Intelligence},
	volume={38},
	number={11},
	pages={12511--12518},
	year={2024}
}

@inproceedings{yan2025ucod,
	title={UCOD-DPL: Unsupervised Camouflaged Object Detection via Dynamic Pseudo-label Learning},
	author={Yan, Weiqi and Chen, Lvhai and Kou, Huaijia and Zhang, Shengchuan and Zhang, Yan and Cao, Liujuan},
	booktitle={Proceedings of the Computer Vision and Pattern Recognition Conference},
	pages={30365--30375},
	year={2025}
}

@inproceedings{gupta2025circod,
	title={CIRCOD: Co-Saliency Inspired Referring Camouflaged Object Discovery},
	author={Gupta, Avi and Jerripothula, Koteswar Rao and Tillo, Tammam},
	booktitle={2025 IEEE/CVF Winter Conference on Applications of Computer Vision (WACV)},
	pages={8313--8323},
	year={2025},
	organization={IEEE}
}

@inproceedings{zhaiexploring,
	title={Exploring Figure-Ground Assignment Mechanism in Perceptual Organization},
	author={Zhai, Wei and Cao, Yang and Zhang, Jing and Zha, Zheng-Jun},
	booktitle={NIPS},
	volume={35},
	year={2022}
}

@inproceedings{He2023Camouflaged,
	title={Camouflaged Object Detection with Feature Decomposition and Edge Reconstruction},
	author={He, Chunming and Li, Kai and Zhang, Yachao and Tang, Longxiang and Zhang, Yulun and Guo, Zhenhua and Li, Xiu},
	booktitle={CVPR},
	year={2023}
}

@article{kirillov2023segment,
	title={Segment anything},
	author={Kirillov, Alexander and Mintun, Eric and Ravi, Nikhila and Mao, Hanzi and Rolland, Chloe and Gustafson, Laura and Xiao, Tete and Whitehead, Spencer and Berg, Alexander C and Lo, Wan-Yen and others},
	journal={arXiv preprint arXiv:2304.02643},
	year={2023}
}

@article{he2022weakly,
	title={Weakly-Supervised Camouflaged Object Detection with Scribble Annotations},
	author={He, Ruozhen and Dong, Qihua and Lin, Jiaying and Lau, Rynson WH},
	journal={AAAI},
	year={2023}
}

@article{skurowski2018animal,
	title={Animal camouflage analysis: Chameleon database},
	author={Skurowski, Przemys{\l}aw and Abdulameer, Hassan and B{\l}aszczyk, J and Depta, Tomasz and Kornacki, Adam and Kozie{\l}, P},
	journal={Unpublished manuscript},
	volume={2},
	number={6},
	pages={7},
	year={2018}
}

@inproceedings{mei2020don,
	title={Don't hit me! glass detection in real-world scenes},
	author={Mei, Haiyang and Yang, Xin and Wang, Yang and Liu, Yuanyuan and He, Shengfeng and Zhang, Qiang and Wei, Xiaopeng and Lau, Rynson WH},
	booktitle={CVPR},
	pages={3687--3696},
	year={2020}
}

@inproceedings{lin2021rich,
	title={Rich context aggregation with reflection prior for glass surface detection},
	author={Lin, Jiaying and He, Zebang and Lau, Rynson WH},
	booktitle={CVPR},
	pages={13415--13424},
	year={2021}
}

@article{kowalski2025bi,
	title={Bi-spectral concealed object detection with attention-based fusion of passive thermal infrared and terahertz imaging},
	author={Kowalski, Marcin and Mierzejewski, Krzysztof and Pa{\l}ys, Tomasz},
	journal={Engineering Applications of Artificial Intelligence},
	volume={158},
	pages={111462},
	year={2025},
	publisher={Elsevier}
}

@article{cheng2025enhancing,
	title={Enhancing concealed object detection in active THz security images with adaptation-YOLO},
	author={Cheng, Aiguo and Wu, Shiyou and Liu, Xiaodong and Lu, Hangyu},
	journal={Scientific Reports},
	volume={15},
	number={1},
	pages={2735},
	year={2025},
	publisher={Nature Publishing Group UK London}
}

@inproceedings{fan2020pranet,
	title={Pranet: Parallel reverse attention network for polyp segmentation},
	author={Fan, Deng-Ping and Ji, Ge-Peng and Zhou, Tao and Chen, Geng and Fu, Huazhu and Shen, Jianbing and Shao, Ling},
	booktitle={MICCAI},
	pages={263--273},
	year={2020},
	organization={Springer}
}

@inproceedings{qin2019basnet,
	title={Basnet: Boundary-aware salient object detection},
	author={Qin, Xuebin and Zhang, Zichen and Huang, Chenyang and Gao, Chao and Dehghan, Masood and Jagersand, Martin},
	booktitle={Proceedings of the IEEE/CVF conference on computer vision and pattern recognition},
	pages={7479--7489},
	year={2019}
}

@inproceedings{wei2020f3net,
	title={F$^3$Net: fusion, feedback and focus for salient object detection},
	author={Wei, Jun and Wang, Shuhui and Huang, Qingming},
	booktitle={Proceedings of the AAAI conference on artificial intelligence},
	volume={34},
	number={07},
	pages={12321--12328},
	year={2020}
}

@inproceedings{zhong2022detecting,
	title={Detecting camouflaged object in frequency domain},
	author={Zhong, Yijie and Li, Bo and Tang, Lv and Kuang, Senyun and Wu, Shuang and Ding, Shouhong},
	booktitle={CVPR},
	pages={4504--4513},
	year={2022}
}

@inproceedings{zhu2022can,
	title={I can find you! Boundary-guided separated attention network for camouflaged object detection},
	author={Zhu, Hongwei and Li, Peng and Xie, Haoran and Yan, Xuefeng and Liang, Dong and Chen, Dapeng and Wei, Mingqiang and Qin, Jing},
	booktitle={AAAI},
	volume={36},
	pages={3608--3616},
	year={2022}
}

@inproceedings{zhang2020adaptive,
	title={Adaptive context selection for polyp segmentation},
	author={Zhang, Ruifei and Li, Guanbin and Li, Zhen and Cui, Shuguang and Qian, Dahong and Yu, Yizhou},
	booktitle={MICCAI},
	pages={253--262},
	year={2020},
	organization={Springer}
}

@inproceedings{xie2020segmenting,
	title={Segmenting transparent objects in the wild},
	author={Xie, Enze and Wang, Wenjia and Wang, Wenhai and Ding, Mingyu and Shen, Chunhua and Luo, Ping},
	booktitle={ECCV},
	year={2020},
	organization={Springer}
}

@article{elsken2019neural,
	title={Neural architecture search: A survey},
	author={Elsken, Thomas and Metzen, Jan Hendrik and Hutter, Frank},
	journal={Journal of Machine Learning Research},
	volume={20},
	number={55},
	pages={1--21},
	year={2019}
}

@article{zoph2016neural,
	title={Neural architecture search with reinforcement learning},
	author={Zoph, Barret and Le, Quoc V},
	journal={arXiv preprint arXiv:1611.01578},
	year={2016}
}

@inproceedings{yao2020sm,
	title={SM-NAS: Structural-to-modular neural architecture search for object detection},
	author={Yao, Lewei and Xu, Hang and Zhang, Wei and Liang, Xiaodan and Li, Zhenguo},
	booktitle={Proceedings of the AAAI conference on artificial intelligence},
	volume={34},
	number={07},
	pages={12661--12668},
	year={2020}
}

@inproceedings{zoph2018learning,
	title={Learning transferable architectures for scalable image recognition},
	author={Zoph, Barret and Vasudevan, Vijay and Shlens, Jonathon and Le, Quoc V},
	booktitle={Proceedings of the IEEE conference on computer vision and pattern recognition},
	pages={8697--8710},
	year={2018}
}

@inproceedings{liu2018progressive,
	title={Progressive neural architecture search},
	author={Liu, Chenxi and Zoph, Barret and Neumann, Maxim and Shlens, Jonathon and Hua, Wei and Li, Li-Jia and Fei-Fei, Li and Yuille, Alan and Huang, Jonathan and Murphy, Kevin},
	booktitle={Proceedings of the European conference on computer vision (ECCV)},
	pages={19--34},
	year={2018}
}

@inproceedings{real2019regularized,
	title={Regularized evolution for image classifier architecture search},
	author={Real, Esteban and Aggarwal, Alok and Huang, Yanping and Le, Quoc V},
	booktitle={Proceedings of the aaai conference on artificial intelligence},
	volume={33},
	number={01},
	pages={4780--4789},
	year={2019}
}

@inproceedings{pham2018efficient,
	title={Efficient neural architecture search via parameters sharing},
	author={Pham, Hieu and Guan, Melody and Zoph, Barret and Le, Quoc and Dean, Jeff},
	booktitle={International conference on machine learning},
	pages={4095--4104},
	year={2018},
	organization={PMLR}
}

@inproceedings{liu2019auto,
	title={Auto-deeplab: Hierarchical neural architecture search for semantic image segmentation},
	author={Liu, Chenxi and Chen, Liang-Chieh and Schroff, Florian and Adam, Hartwig and Hua, Wei and Yuille, Alan L and Fei-Fei, Li},
	booktitle={Proceedings of the IEEE/CVF conference on computer vision and pattern recognition},
	pages={82--92},
	year={2019}
}

@inproceedings{zhang2019customizable,
	title={Customizable architecture search for semantic segmentation},
	author={Zhang, Yiheng and Qiu, Zhaofan and Liu, Jingen and Yao, Ting and Liu, Dong and Mei, Tao},
	booktitle={Proceedings of the IEEE/CVF Conference on Computer Vision and Pattern Recognition},
	pages={11641--11650},
	year={2019}
}

@inproceedings{nekrasov2019fast,
	title={Fast neural architecture search of compact semantic segmentation models via auxiliary cells},
	author={Nekrasov, Vladimir and Chen, Hao and Shen, Chunhua and Reid, Ian},
	booktitle={Proceedings of the IEEE/CVF conference on computer vision and pattern recognition},
	pages={9126--9135},
	year={2019}
}

@inproceedings{lin2020graph,
	title={Graph-guided architecture search for real-time semantic segmentation},
	author={Lin, Peiwen and Sun, Peng and Cheng, Guangliang and Xie, Sirui and Li, Xi and Shi, Jianping},
	booktitle={Proceedings of the IEEE/CVF conference on computer vision and pattern recognition},
	pages={4203--4212},
	year={2020}
}

@article{chen2019detnas,
	title={Detnas: Backbone search for object detection},
	author={Chen, Yukang and Yang, Tong and Zhang, Xiangyu and Meng, Gaofeng and Xiao, Xinyu and Sun, Jian},
	journal={Advances in neural information processing systems},
	volume={32},
	year={2019}
}

@inproceedings{guo2020hit,
	title={Hit-detector: Hierarchical trinity architecture search for object detection},
	author={Guo, Jianyuan and Han, Kai and Wang, Yunhe and Zhang, Chao and Yang, Zhaohui and Wu, Han and Chen, Xinghao and Xu, Chang},
	booktitle={Proceedings of the IEEE/CVF conference on computer vision and pattern recognition},
	pages={11405--11414},
	year={2020}
}

@inproceedings{wang2020fcos,
	title={NAS-FCOS: Fast neural architecture search for object detection},
	author={Wang, Ning and Gao, Yang and Chen, Hao and Wang, Peng and Tian, Zhi and Shen, Chunhua and Zhang, Yanning},
	booktitle={proceedings of the IEEE/CVF conference on computer vision and pattern recognition},
	pages={11943--11951},
	year={2020}
}

@inproceedings{real2017large,
	title={Large-scale evolution of image classifiers},
	author={Real, Esteban and Moore, Sherry and Selle, Andrew and Saxena, Saurabh and Suematsu, Yutaka Leon and Tan, Jie and Le, Quoc V and Kurakin, Alexey},
	booktitle={International conference on machine learning},
	pages={2902--2911},
	year={2017},
	organization={PMLR}
}

@article{xu2019pc,
	title={Pc-darts: Partial channel connections for memory-efficient architecture search},
	author={Xu, Yuhui and Xie, Lingxi and Zhang, Xiaopeng and Chen, Xin and Qi, Guo-Jun and Tian, Qi and Xiong, Hongkai},
	journal={arXiv preprint arXiv:1907.05737},
	year={2019}
}

@inproceedings{chen2019progressive,
	title={Progressive differentiable architecture search: Bridging the depth gap between search and evaluation},
	author={Chen, Xin and Xie, Lingxi and Wu, Jun and Tian, Qi},
	booktitle={Proceedings of the IEEE/CVF international conference on computer vision},
	pages={1294--1303},
	year={2019}
}

@article{chu2020darts,
	title={Darts-: robustly stepping out of performance collapse without indicators},
	author={Chu, Xiangxiang and Wang, Xiaoxing and Zhang, Bo and Lu, Shun and Wei, Xiaolin and Yan, Junchi},
	journal={arXiv preprint arXiv:2009.01027},
	year={2020}
}

@article{cheng2020hierarchical,
	title={Hierarchical neural architecture search for deep stereo matching},
	author={Cheng, Xuelian and Zhong, Yiran and Harandi, Mehrtash and Dai, Yuchao and Chang, Xiaojun and Li, Hongdong and Drummond, Tom and Ge, Zongyuan},
	journal={Advances in neural information processing systems},
	volume={33},
	pages={22158--22169},
	year={2020}
}

@inproceedings{na2022autosnn,
	title={Autosnn: Towards energy-efficient spiking neural networks},
	author={Na, Byunggook and Mok, Jisoo and Park, Seongsik and Lee, Dongjin and Choe, Hyeokjun and Yoon, Sungroh},
	booktitle={International conference on machine learning},
	pages={16253--16269},
	year={2022},
	organization={PMLR}
}

@inproceedings{ghiasi2019fpn,
	title={Nas-fpn: Learning scalable feature pyramid architecture for object detection},
	author={Ghiasi, Golnaz and Lin, Tsung-Yi and Le, Quoc V},
	booktitle={Proceedings of the IEEE/CVF conference on computer vision and pattern recognition},
	pages={7036--7045},
	year={2019}
}

@inproceedings{tan2019mnasnet,
	title={Mnasnet: Platform-aware neural architecture search for mobile},
	author={Tan, Mingxing and Chen, Bo and Pang, Ruoming and Vasudevan, Vijay and Sandler, Mark and Howard, Andrew and Le, Quoc V},
	booktitle={Proceedings of the IEEE/CVF conference on computer vision and pattern recognition},
	pages={2820--2828},
	year={2019}
}

@inproceedings{tan2020efficientdet,
	title={Efficientdet: Scalable and efficient object detection},
	author={Tan, Mingxing and Pang, Ruoming and Le, Quoc V},
	booktitle={Proceedings of the IEEE/CVF conference on computer vision and pattern recognition},
	pages={10781--10790},
	year={2020}
}

@article{cai2018proxylessnas,
	title={Proxylessnas: Direct neural architecture search on target task and hardware},
	author={Cai, Han and Zhu, Ligeng and Han, Song},
	journal={arXiv preprint arXiv:1812.00332},
	year={2018}
}

@inproceedings{huang2023adaptive,
	title={Adaptive frequency filters as efficient global token mixers},
	author={Huang, Zhipeng and Zhang, Zhizheng and Lan, Cuiling and Zha, Zheng-Jun and Lu, Yan and Guo, Baining},
	booktitle={Proceedings of the IEEE/CVF international conference on computer vision},
	pages={6049--6059},
	year={2023}
}

@article{he2025frequency,
	title={Frequency-Domain Fusion Transformer for Image Inpainting},
	author={He, Sijin and Lin, Guangfeng and Li, Tao and Chen, Yajun},
	journal={arXiv preprint arXiv:2506.18437},
	year={2025}
}

@article{geng2021attention,
	title={Is attention better than matrix decomposition?},
	author={Geng, Zhengyang and Guo, Meng-Hao and Chen, Hongxu and Li, Xia and Wei, Ke and Lin, Zhouchen},
	journal={arXiv preprint arXiv:2109.04553},
	year={2021}
}

@inproceedings{li2021uncertainty,
	title={Uncertainty-aware joint salient object and camouflaged object detection},
	author={Li, Aixuan and Zhang, Jing and Lv, Yunqiu and Liu, Bowen and Zhang, Tong and Dai, Yuchao},
	booktitle={Proceedings of the IEEE/CVF conference on computer vision and pattern recognition},
	pages={10071--10081},
	year={2021}
}

@article{yue2025cnn,
	title={When CNN meet with ViT: decision-level feature fusion for camouflaged object detection},
	author={Yue, Guowen and Jiao, Ge and Li, Chen and Xiang, Jiahao},
	journal={The Visual Computer},
	volume={41},
	number={6},
	pages={3957--3972},
	year={2025},
	publisher={Springer}
}

@article{ge2025feature,
	title={Feature-aware and iterative refinement network for camouflaged object detection},
	author={Ge, Yanliang and Ren, Junchao and Zhang, Cong and He, Min and Bi, Hongbo and Zhang, Qiao},
	journal={The Visual Computer},
	volume={41},
	number={7},
	pages={4741--4758},
	year={2025},
	publisher={Springer}
}

@article{Zhang2023MutualAttention,
	author  = {Zhang, Mengting and Tian, Xiuxia},
	title   = {Transformer architecture based on mutual attention for image-anomaly detection},
	journal = {Virtual Reality \& Intelligent Hardware},
	volume  = {5},
	number  = {1},
	pages   = {57--67},
	year    = {2023},
	doi     = {10.1016/j.vrih.2022.07.006}
}

@article{Lin2025SCNet,
	author  = {Lin, Chuchao and Zou, Changjun and Xu, Hangbin},
	title   = {SCNet: A Dual-Branch Network for Strong Noisy Image Denoising Based on Swin Transformer and ConvNeXt},
	journal = {Computer Animation and Virtual Worlds},
	volume  = {36},
	number  = {3},
	pages   = {e70030},
	year    = {2025},
	doi     = {10.1002/cav.70030}
}

@article{Zhang2025JointLearning,
	author  = {Zhang, Mengyao and Zhou, Jie and Miao, Tingyun and Zhao, Yong and Si, Xin and Zhang, Jingliang},
	title   = {Joint-Learning: A Robust Segmentation Method for 3D Point Clouds Under Label Noise},
	journal = {Computer Animation and Virtual Worlds},
	volume  = {36},
	number  = {3},
	pages   = {e70038},
	year    = {2025},
	doi     = {10.1002/cav.70038}
}

@article{Chen2023MNGNAS,
	author  = {Chen, Zhihua and Qiu, Guhao and Li, Ping and Zhu, Lei and Yang, Xiaokang and Sheng, Bin},
	title   = {MNGNAS: Distilling Adaptive Combination of Multiple Searched Networks for One-Shot Neural Architecture Search},
	journal = {IEEE Transactions on Pattern Analysis and Machine Intelligence},
	volume  = {45},
	number  = {11},
	pages   = {13489--13508},
	year    = {2023},
	doi     = {10.1109/TPAMI.2023.3293885}
}

@article{Lin2023EAPT,
	author  = {Lin, Xiao and Sun, Shuzhou and Huang, Wei and Sheng, Bin and Li, Ping and Feng, David Dagan},
	title   = {EAPT: Efficient Attention Pyramid Transformer for Image Processing},
	journal = {IEEE Transactions on Multimedia},
	volume  = {25},
	pages   = {50--61},
	year    = {2023},
	doi     = {10.1109/TMM.2021.3120873}
}

@inproceedings{li2025camouflaged,
	title={Camouflaged Object Detection via Neural Architecture Search},
	author={Li, Xin and Fu, Keren and Zhao, Qijun},
	booktitle={ICASSP 2025-2025 IEEE International Conference on Acoustics, Speech and Signal Processing (ICASSP)},
	pages={1--5},
	year={2025},
	organization={IEEE}
}

@article{pang2024zoomnext,
	title={Zoomnext: A unified collaborative pyramid network for camouflaged object detection},
	author={Pang, Youwei and Zhao, Xiaoqi and Xiang, Tian-Zhu and Zhang, Lihe and Lu, Huchuan},
	journal={IEEE transactions on pattern analysis and machine intelligence},
	volume={46},
	number={12},
	pages={9205--9220},
	year={2024},
	publisher={IEEE}
}

\end{document}